\documentclass[letterpaper, 10 pt, conference]{ieeeconf}  

\IEEEoverridecommandlockouts                              

\usepackage{graphics} 
\usepackage{epsfig} 
\usepackage{mathptmx} 
\usepackage{subfloat}
\usepackage{float}
\usepackage{wrapfig}
\usepackage{subfig}
\usepackage{amsmath} 
\title{\LARGE \bf
A deep representation for depth images from synthetic data
}

\author{Fabio Maria Carlucci$^{1}$ and Paolo Russo$^{1}$ and Barbara Caputo$^{1}$
\thanks{*This work was supported by the projects ALOOF CHIS-ERA (F.M.C.) and ERC RoboExNovo (B.C.).We thank S. Baharlou, A. Gigli, F. Giordaniello, M. Graziani and M. Lampacrescia for their help in creating the VANDAL database.}
\thanks{$^{1}$F. M. Carlucci, P. Russo and B. Caputo are with the VANDAL Laboratory, Department of Computer, Management and Control Engineering (DIAG), Sapienza Rome University, Rome, Italy 
        {\tt\small fabiom.carlucci@dis.uniroma1.it}}%
}

\begin{document}

\maketitle
\thispagestyle{empty}
\pagestyle{empty}

\begin{abstract}
Convolutional Neural Networks (CNNs) trained on large scale RGB databases have become the secret sauce in the majority of recent approaches for object categorization from RGB-D data.
Thanks to colorization techniques, these methods exploit the filters learned from 2D images to extract meaningful representations in 2.5D. Still, the perceptual signature of these two kind of images is very different, with the first usually strongly  characterized by textures, and the second mostly by silhouettes of objects. Ideally, one would like to have two CNNs, one for RGB and one for depth, each trained on a suitable data collection, able to capture the perceptual properties of each channel for the task at hand. This has not been possible so far, due to the lack of a suitable depth database. This paper addresses this issue, proposing to opt for synthetically generated images rather than collecting by hand a 2.5D large scale database. While being clearly a proxy for real data, synthetic images allow to trade quality for quantity, making it possible to generate a virtually infinite amount of data. We show that the filters learned from such data collection, using the very same architecture typically used on visual data, learns very different filters, resulting in depth features (a)  able to better characterize the different facets of depth images, and (b) complementary with respect to those derived from CNNs pre-trained on 2D datasets. Experiments on two publicly available databases show the power of our approach.

\end{abstract}

\section{INTRODUCTION}

Deep learning has changed the research landscape in visual object recognition over the last few years. Since their spectacular success in recognizing $1,000$ object categories \cite{alexnet}, convolutional neural networks have become the new off the shelf state of the art in visual classification. Since then, the robot vision community has also attempted to 
take advantage of the deep learning trend, as the ability of robots to understand what they see reliably is critical for their deployment in the wild. A critical issue when trying to transfer results from computer to robot vision is that robot perception is tightly coupled with robot action. Hence, pure RGB visual recognition is not enough.  

The heavy use of 2.5D depth sensors on robot platforms has generated a lively research activity on 2.5D object recognition from depth maps \cite{spin,rcnn,Cheng}. Here a strong emerging trend is that of using Convolutional Neural Networks (CNNs) pre-trained over ImageNet \cite{imagenet} by colorizing the depth channel \cite{Behnke}.  The approach has proved successful, especially when coupled with fine tuning \cite{Eitel} and/or spatial pooling strategies \cite{hasan,ChengZhao2014,ChengZhao2015} (for a review of recent work we refer to section \ref{rel-work}). These results suggest that the filters learned by CNNs from ImageNet are able to capture information also from depth images, regardless of their perceptual difference. 

Is this the best we can do? What if  one would train from scratch a CNN over a very large scale 2.5D object categorization database, wouldn't the filters learned be more suitable for object recognition from depth images? RGB images are perceptually very rich, with generally a strong presence of textured patterns, especially in ImageNet. Features learned from RGB data are most likely focusing on those aspects, while depth images contain more information about the shape and the silhouette of objects.
Unfortunately, as of today a 2.5D object categorization database large enough to train a CNN on it does not exist. A likely reason for this is that gathering such data collection is a daunting challenge: capturing the same variability of ImageNet over the same number of object categories would require the coordination of very many laboratories, over an extended period of time.

\begin{figure*}[!htb]
\centering
\subfloat []{%
 \includegraphics[width=0.24\textwidth]{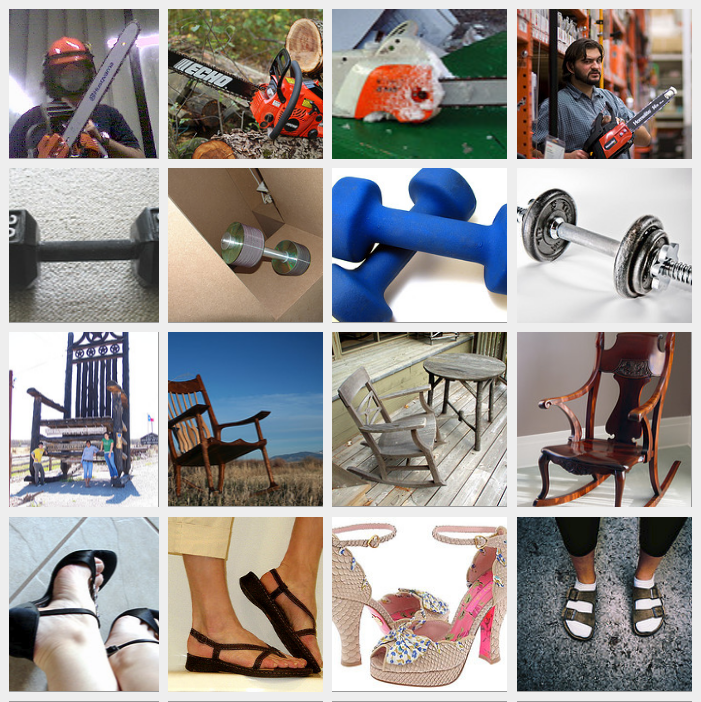}
}
\subfloat []{%
 \includegraphics[width=0.24\textwidth]{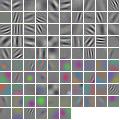}
}
\subfloat[] {%
\includegraphics[width=0.24\textwidth]{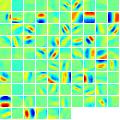}
}
\subfloat []{%
 \includegraphics[width=0.24\textwidth]{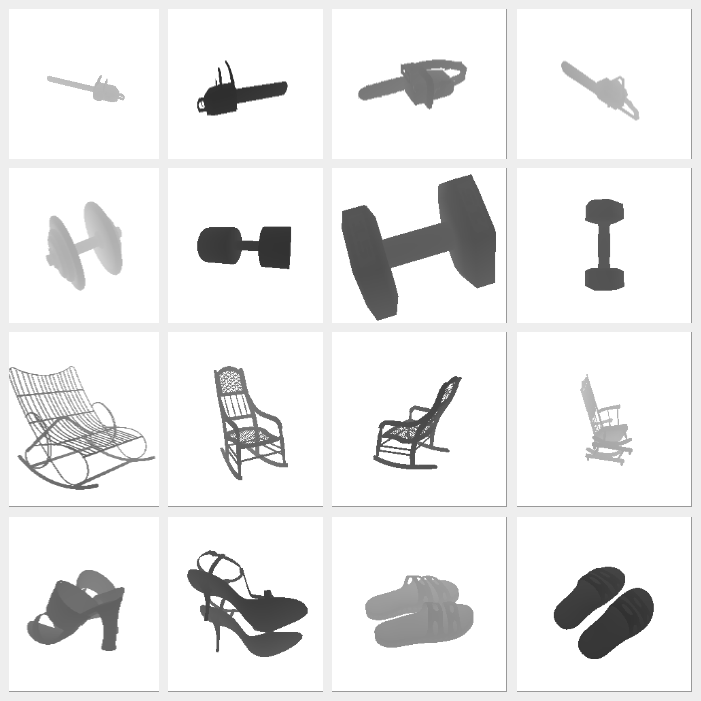}
}
\caption{Sample images for the classes chainsaw, dumbbell, rocker chair and sandal from ImageNet (a) and VANDAL (d). We show the corresponding filters learned by the very same CNN architecture respectively in (b) and (c). We see that even though the architecture is the same, using 2D rather than 2.5D images for training leads to learning quite different filters.}
\label{fig:first_layer_weights}
\end{figure*}

In this paper we follow an alternative route. Rather than acquiring a 2.5D object categorization database, we propose to use synthetic data as a proxy for training a deep learning architecture specialized in learning depth specific features. To this end, we construct the VANDAL database, a collection of $4.1$ million depth images from more than $9,000$ objects, belonging to $319$ categories. The depth images are generated starting from 3D CAD models, downloaded from the Web, through a protocol developed to extract the maximum information from the models. VANDAL is used as input to train from scratch a deep learning architecture, obtaining a pre-trained model able to act as a depth specific feature extractor. Visualizations  of the filters learned by the first layer of the architecture show that the filter we obtain are indeed very different from those learned from ImageNet with the very same convolutional neural network (figure \ref{fig:first_layer_weights}). As such, they are able to capture different facets of the perceptual information available from real depth images, more suitable for the recognition task in that domain. We call our pre-trained architecture DepthNet. 

Experimental results on two publicly available databases confirm this: when using only depth, our DepthNet features achieve better performance compared to previous methods based on a CNN pre-trained over ImageNet, without using fine tuning or spatial pooling. The combination of the DepthNet features with the descriptors obtained from the CNN pre-trained over ImageNet, on both depth and RGB images, leads to strong results on the Washington database \cite{washington}, and to results competitive with fine-tuning and/or sophisticated spatial pooling approaches on the JHUIT database \cite{jhuit}. To the best of our knowledge, this is the first work that uses synthetically generated depth data to train a depth-specific convolutional neural network. Upon acceptance of the paper, all the VANDAL data, the protocol and the software for generating new depth images, as well as the pre-trained DepthNet, will be made publicly available.

The rest of the paper is organized as follows. After a review of the recent literature (section \ref{rel-work}), we introduce the VANDAL database, describing its generation protocol and showcasing the obtained depth images (section \ref{db}). Section \ref{net} describes the deep architecture used and section \ref{expers} reports our experimental findings. The paper concludes with a summary and a discussion on future research.

\section{RELATED WORKS}
\label{rel-work}

Object recognition from RGB-D data traditionally relied on hand-crafted features such as SIFT \cite{sift} and spin images \cite{spin}, combined together through vector quantization in a Bag-of-Words encoding \cite{spin}. This heuristic approach has been surpassed by end-to-end feature learning architectures, able to define suitable features in a data-driven fashion   \cite{ckm,rcnn,crf}. All these methods have been designed to cope with a limited amount of training data (of the order of $10^3-10^4$ depth images), thus they are able to only partially exploit the generalization abilities of deep learning as feature extractors experienced in the computer vision community \cite{alexnet,razavian}, where databases of $10^6$ RGB images like ImageNet \cite{imagenet} or Places \cite{places} are available.

An alternative route is that of re-using deep learning architectures trained on ImageNet through pre-defined encoding \cite{gupta} or colorization. Since the work of \cite{Behnke} re-defined the state of the art in the field, this last approach has been actively and successfully investigated.
Eitel et al \cite{Eitel} proposed a parallel CNN architecture, one for the depth channel and one for the RGB one, combined together in the final layers through a late fusion scheme. 
Some approaches coupled non linear learning methods with various forms of spatial encodings \cite{ChengZhao2015,ChengZhao2014,Cheng,jhuit}.
Hasan et al \cite{hasan} pushed further this multi-modal approach, proposing an architecture merging together RGB, depth and 3D point cloud information. Another notable feature is the encoding of an implicit multi scale representation through a rich coarse-to-fine feature extraction approach. 

All these works build on top of CNNs pre-trained over ImageNet, for all modal channels. Thus, the very same filters are used to extract features from all of them. As empirically successful as this might be, it is a questionable strategy, as RGB and depth images are perceptually very different, and as such they would benefit from approaches able to learn data-specific features (figure \ref{fig:first_layer_weights}). Our method matches this challenge, learning RGB features from RGB data and depth features from synthetically generated data, within a deep learning framework. The use of realistic synthetic data in conjunction with deep learning architectures is a promising emerging trend \cite{cvpr15-papon,voxnet,shapenet}. We are not aware of previous work attempting to use synthetic data to learn depth representations, with or without deep learning techniques.

\section{THE VANDAL DATABASE}
\label{db}
In this section we present VANDAL and the protocol followed for its creation. With $4,106,340$ synthetic images, it is the largest existing depth database for object recognition. Section \ref{db-1} describes the criteria used to select the object categories composing the database and the protocol followed to obtain the 3D CAD models from Web resources. Section \ref{db-2} illustrates the procedure used to generate depth images from the 3D CAD models. 

\begin{figure}[!htb]
\centering
\subfloat {
 \includegraphics[width=0.5\textwidth]{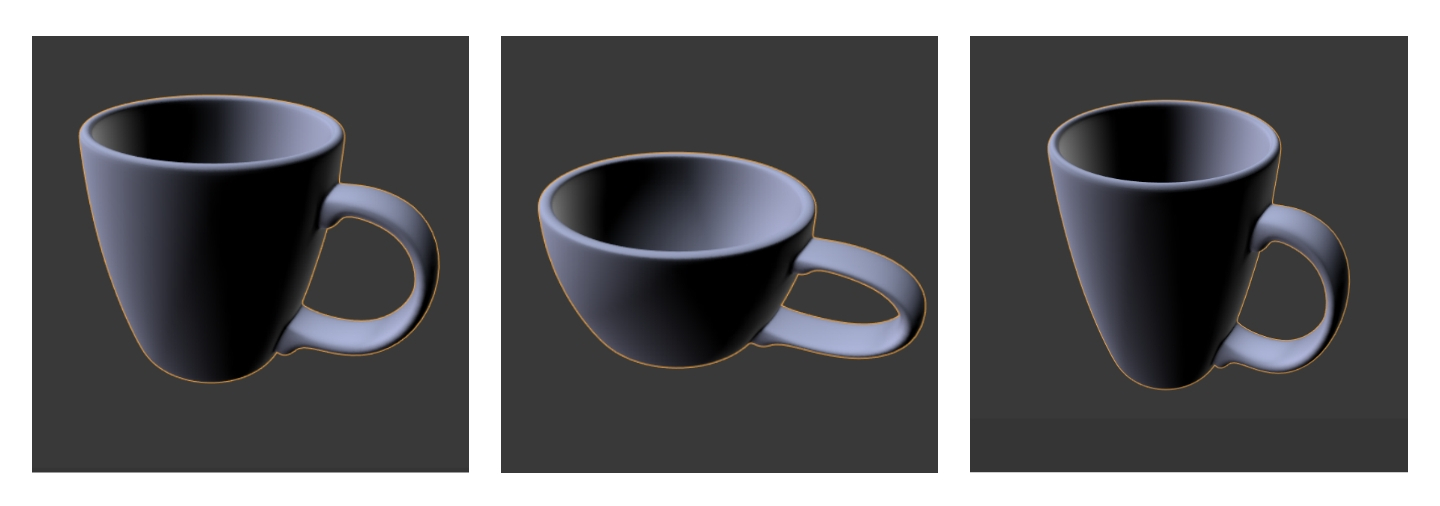}
}
\caption{Sample morphs (center, right) generated from an instance model for the category coffee cup (left). }
\label{fig:morphs}
\end{figure}
\begin{figure*}[!htb]
\centering
\subfloat [The 319 categories in VANDAL] {
\includegraphics[width=0.55\textwidth,height=0.25\textheight]{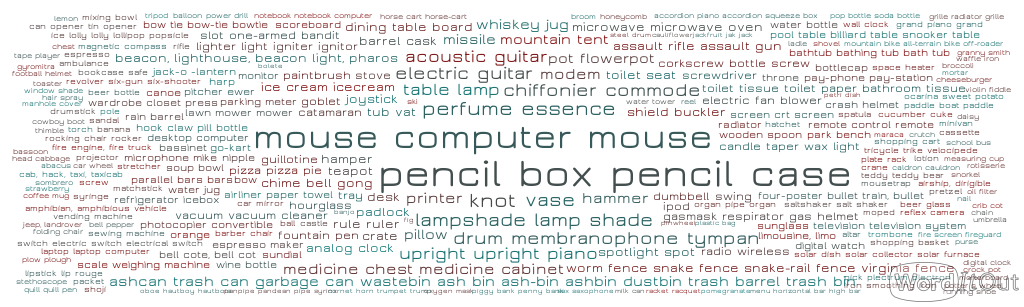}
}
\subfloat [Some examples from the classes more populated]{
\includegraphics[width=0.4\textwidth,height=0.25\textheight]{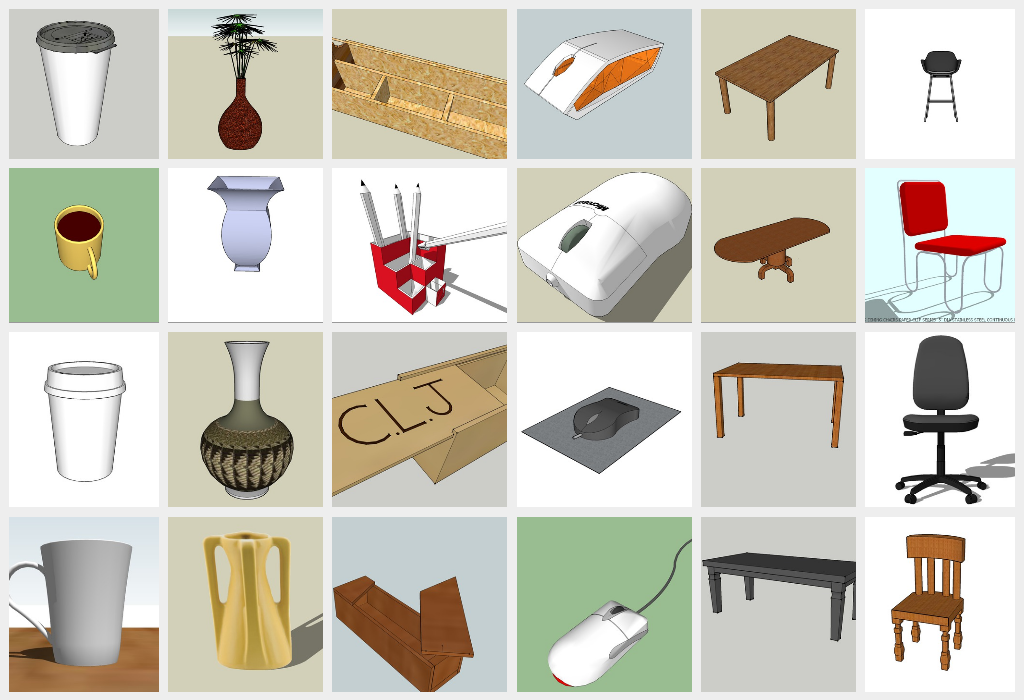}
}
\caption{The VANDAL database. On the left, we show a word cloud visualization of the classes in it, based on the numbers of 3D models in each category. On the right, we show exemplar models for the six categories more populated: coffee cup, vase, pencil case, computer mouse, table and chair.}
\label{fig:Vandal}
\end{figure*}
\subsection{Selecting and Generating the 3D Models} 
\label{db-1}
CNNs trained on ImageNet have been shown to generalize well when used on other object centric datasets. Following this reasoning, we defined a list of object categories as a subset of the ILSVRC2014 list \cite{imagenet}, removing by hand all scenery classes, as well as objects without a clear default shape such as clothing items or animals. This resulted in a first list of roughly 480 categories, which was used to query public 3D CAD model repositories like 3D Warehouse, Yeggi, Archive3D, and many others.
Five volunteers\footnote{Graduate students from the MARR program at DIAG, Sapienza Rome University.} manually downloaded the models, removing all irrelevant items like floor or other supporting surfaces, people standing next to the object and so forth, and running a script to harmonize the size of all models (some of them were originally over 1GB per file). They were also required to create significantly morphed variations of the original 3D CAD models, whenever suitable. 
Figure \ref{fig:morphs} shows examples of morphed models for the object category coffee cup. 
Finally, we removed all categories with less than two models, ending up with 319 object categories with an average of 30 models per category, for a total of $9,383$ CAD object models. Figure \ref{fig:Vandal}, left, gives a world cloud visualization of the VANDAL dataset, while on the right it shows examples of 3D models for the 6 most populated object categories.





\subsection{From 3D Models to 2.5 Depth Images} 
\label{db-2}
All depth renderings were created using Blender\footnote{\texttt{www.blender.org}}, with a python script fully automating the procedure, and then saved as grayscale .png files, using the convention that black is close and white is far. 

The depth data generation protocol was designed to extract as much information as possible from the available 3D CAD models. This concretely means obtaining the greatest possible variability between each rendering. The approach commonly used by real RGB-D datasets consists in fixing the camera at a given angle and then using a turntable to get all possible viewpoints of the object \cite{washington,jhuit}. We tested here a similar approach, but we found out using perceptual hashing that a significant number of object categories had more than 50\% nearly identical images.

\begin{figure}[!b]
\centering
\includegraphics[width=0.43\textwidth,height=0.26\textheight]{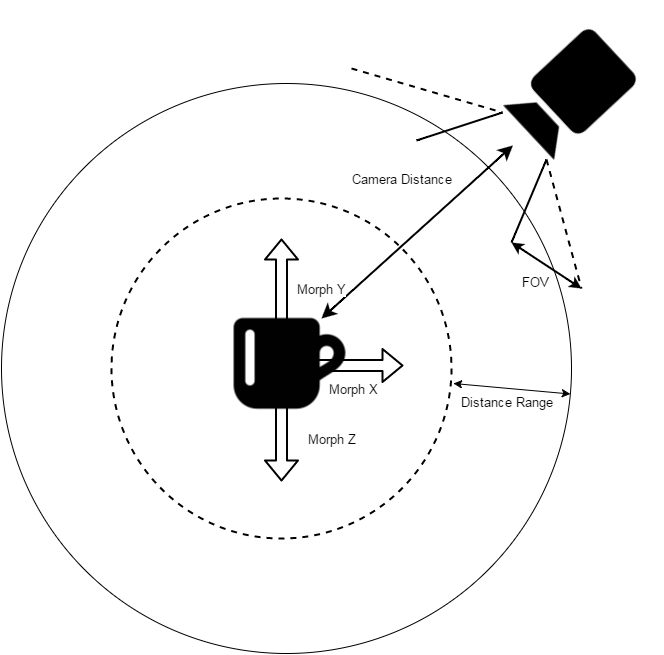}
\caption{Configuration space used for generating renderings in the VANDAL database.}
\label{fig:rendering}
\end{figure}

We defined instead a configuration space consisting of: 
(a) object distance from the camera,
(b) focal length of the camera,
(c) camera position on the sphere defined by the distance, and
 (d) slight (\(< 10\%\)) random morphs along the axes of the model.
 Figure \ref{fig:rendering}  illustrates the described configuration space. This protocol 
ensured that almost none of the resulting images were 
identical.
We sampled this configuration space with roughly $480$ depth images for each model, obtaining a total of $4.1$ million images. Preliminary experiments showed that increasing the sampling rate in the configuration space did lead to growing percentages of nearly identical images.

\begin{figure}[!tb]
\centering
\subfloat {%
 \includegraphics[width=0.48\textwidth]{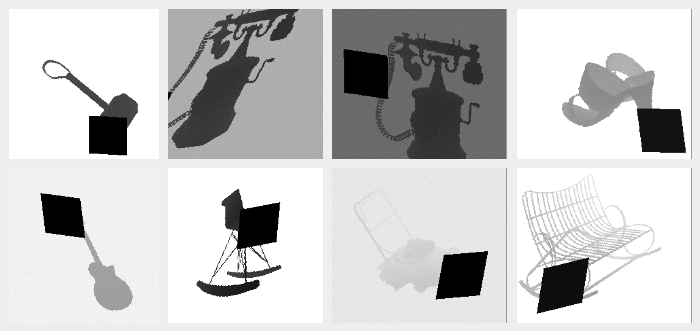}
}
\caption{Data augmentation samples from various classes (hammer, phone, sandal, guitar, rocker, lawn mower, bench). Note that the contrast brightness variations and noise are hard to visualize on small thumbnails.}
\label{fig:data_aug}
\end{figure}

The rendered depth images consist of objects always centered on a white background.
This is done on purpose, as it allows us the maximum freedom to perform various types of data augmentation at training time, as it is standard practice when training convolutional neural networks. This is here even more relevant than usual, as synthetically generated data are intrinsically perceptually less informative compared to real data. 
The data augmentation methods we used are:
image cropping,
occlusion (1/4 of the image is randomly occluded to simulate gaps in the sensor scan),
contrast/brightness variations, in depth views corresponding to scaling the Z axis and shifting the objects along it,
background substitution (substituting the white background with one randomly chosen farther away than the object's center of mass),
 random uniform noise (as in film grain),
and image shearing (a slanting transform). Figure \ref{fig:data_aug} shows some examples of data augmentation images obtained with this protocol.



\section{LEARNING DEEP DEPTH FILTERS}
\label{net}
Once the VANDAL database has been generated, it is possible to use it to train any kind of convolutional deep architecture. In order to allow for a fair comparison with previous work, we opted for CaffeNet, a slight variation of AlexNet \cite{alexnet}. Although more modern networks have been proposed in the last years \cite{vgg,inception,resnet}, it still represents the most popular choice among practitioners, and the most used in robot vision\footnote{Preliminary experiments using the VGG, Inception and Wide Residual networks on the VANDAL database did not give stable results and need further investigation.}.  Its well know architecture consists of 5 convolutional layers, interwoven with pooling, normalization and relu layers, plus three fully connected layers. CaffeNet differs from AlexNet in the pooling, which is done there before normalization. It usually performs slightly better and has thus gained wide popularity.

Although the standard choice in robot vision is using the output of the seventh activation layer as feature descriptors, several studies in the vision community show that lower layers, like the sixth and the fifth, tend to have higher generalization properties \cite{CNNTransfer}. We followed this trend, and opted for the fifth layer (by vectorization) as deep depth feature descriptor
(an ablation study supporting this choice is reported in section \ref{expers}). We name in the following as \textbf{DepthNet} the CaffeNet architecture trained on VANDAL using as output feature the fifth layer, and \textbf{Caffe-ImageNet} the same architecture trained over ImageNet.

\begin{figure*}[!t]
\centering

\subfloat {%
\includegraphics[width=1.0\textwidth]{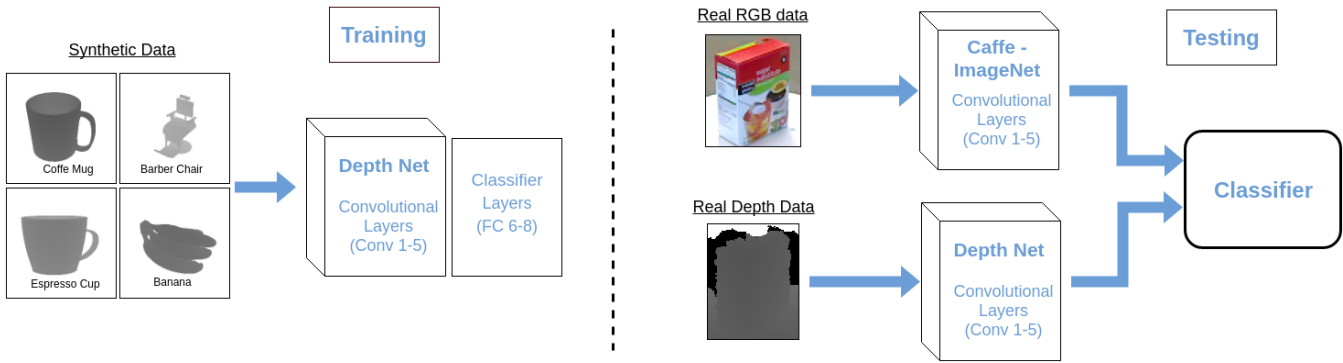}
}
\caption{DepthNet and our associated RGB-D object classification framework. During training, we learn depth filters from the VANDAL synthetic data (left). During test (right), real RGB and depth data is processed by two distinct CNNs, each specialized over the corresponding modality. The features, derived from the activations of the fifth convolutional layer, are then fed into a cue integration classifier.}
\label{fig:architecture}
\end{figure*}

Once DepthNet has been trained, it can be used as any depth feature descriptor, alone or in conjunction with Caffe-ImageNet for classification of RGB images. We explore this last option, proposing a system for RGB-D object categorization that combines the two feature representations through a multi kernel learning classifier \cite{obscure}. Figure \ref{fig:architecture} gives an overview of the overall RGB-D classification system. Note that DepthNet can be combined with any other RGB and/or 3D point cloud descriptor, and that the integration of the modal representations can be achieved through any other cue integration approach. This underlines the versatility of DepthNet, as opposed to recent work where the depth component was tightly integrated within the proposed overall framework, and as such unusable outside of it \cite{Eitel,hasan,Cheng,jhuit}. 

\section{EXPERIMENTS}
\label{expers}
We assessed the DepthNet, as well as the associated RGB-D framework of figure \ref{fig:architecture}, on two publicly available databases. Section \ref{setup} describes our experimental setup and the databases used in our experiments. Section \ref{ablation} reports a set of experiments assessing the performance of DepthNet on depth images, compared to Caffe-ImageNet, while in section \ref{rgb-d} we assess the performance of the whole RGB-D framework with respect to previous approaches.

\subsection{Experimental setup}
\label{setup}
We conducted experiments on the Washington RGB-D \cite{washington} and the JHUIT-50 \cite{jhuit}  object datasets. The first consists of $41,877$ RGB-D images organized into $300$ instances divided in $51$ classes. Each object instance was positioned on a turntable and captured from three different viewpoints while rotating. Since two consecutive views are extremely similar, only 1 frame out of 5 is used for evaluation purposes. We performed experiments on the object categorization setting, where we followed the evaluation protocol defined in \cite{washington}. The second is a challenging recent dataset that focuses on the problem of fine-grained recognition. It contains $50$ object instances, often very similar with each other (e.g. 9 different kinds of screwdrivers). As such, it presents different classification challenges compared to the Washington database. 

All experiments, as well as the training of DepthNet, were done using the publicly available Caffe framework \cite{caffe}, together with NVIDIA Deep Learning GPU Training System (DIGITS). As described above, we obtained DepthNet by training a CaffeNet over the VANDAL database. The network was trained using Stochastic Gradient Descent for 50 epochs. Learning rate started at 0.01 and gamma at 0.5 (halving the learning rate at each step). We used a variable step down policy, where the first step took 25 epochs, the next 25/2, the third 25/4 epochs and so on. These parameters were chosen to make sure that the test loss on the VANDAL test data had stabilized at each learning rate. Weight decay and momentum were left at their standard values of 0.0005 and 0.9. 


To assess the quality of the DepthNet features we performed three set of experiments:
\begin{enumerate}
\item \emph{Object classification using depth only:}  features were extracted with DepthNet and a linear SVM\footnote{Liblinear: http://www.csie.ntu.edu.tw/~cjlin/liblinear/} was trained on it. We also examined how the performance varies when extracting from different layers of the network, comparing against a Caffe-ImageNet used for depth classification, as in \cite{Behnke}.
\item \emph{Object classification using RGB + Depth:}  in this setting we combined our depth features with those extracted from the RGB images using  Caffe-ImageNet. While \cite{Eitel} train a fusion network to do this, we simply use an off the shelf Multi Kernel Learning (MKL) classifier \cite{obscure}. 
\end{enumerate} 
For all experiments we used the training/testing splits originally proposed for each given dataset. For linear SVM, we set $C$ by cross validation. When using MKL, we left the default values of $100$ iterations for online and $300$ for batch and set $p$ and $C$ by cross validation.

Previous works using  Caffe-ImageNet as feature extractor for depth, apply some kind of input preproccessing \cite{Behnke,Eitel,hasan}. While we do compare against the published baselines, we also found that by simply normalizing each image (min to 0 and max to 255), one achieves very competitive results.
Also, since our DepthNet is trained on depth data, it does not need any type of preprocessing over the depth images, obtaining strong results over raw data.
Because of this, in all experiments reported in the following we only consider raw depth images and normalized depth images.

\subsection{Assessing the performance of the DepthNet architecture}
\label{ablation}

\begin{figure}[!b]
\centering
\subfloat {
 \includegraphics[width=0.5\textwidth]{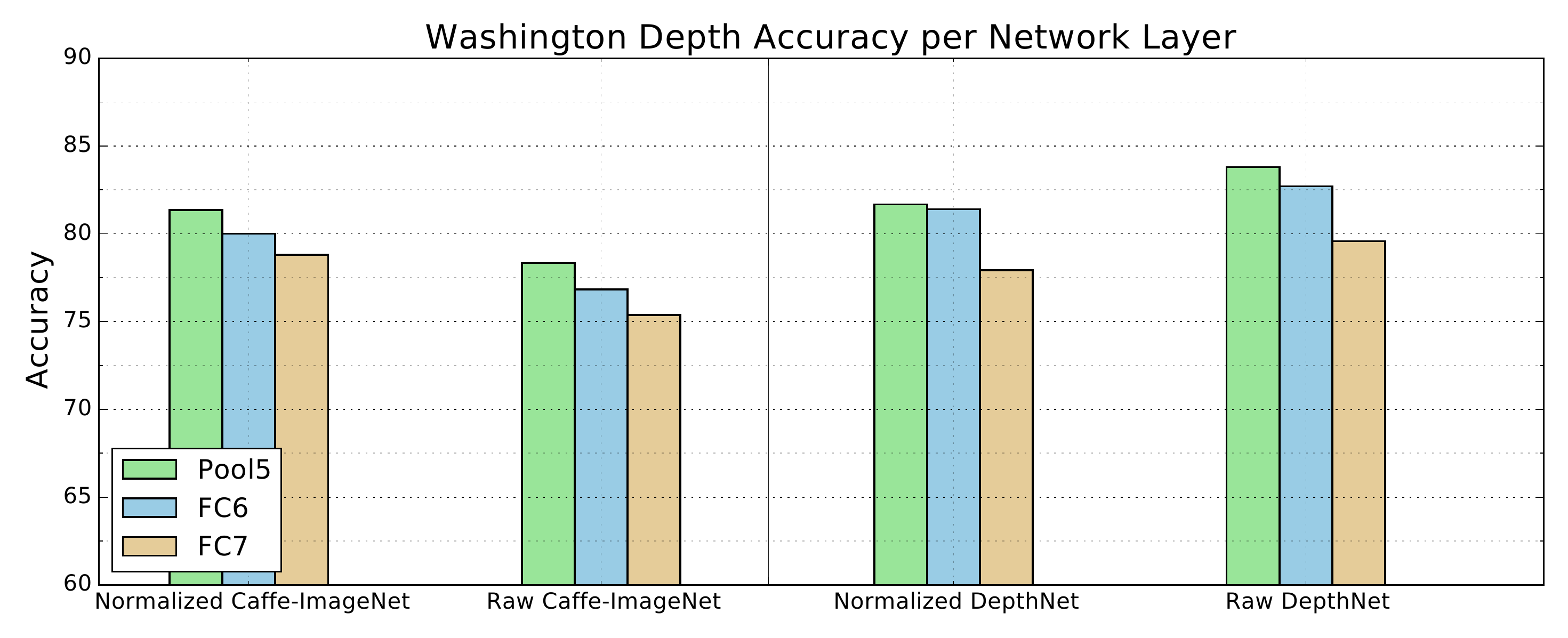}
}
\caption{Accuracy obtained by DepthNet and Caffe-ImageNet over the Washington database, using as features pool5, FC6 and FC7. Results are reported for raw and normalized depth images.}
\label{fig:washington_ablation}
\end{figure}

We present here an ablation study, aiming at understanding the impact of choosing features from the last fully convolutional layer as opposed to the more popular last fully connected layer, and of using normalized depth images instead of raw data. By comparing our results with those obtained by  Caffe-ImageNet, we also aim at illustrating up to which point the features learned from VANDAL are different from those derived from ImageNet.
\begin{figure*}[!tb]
\centering
\subfloat {
 \includegraphics[width=1.0\textwidth]{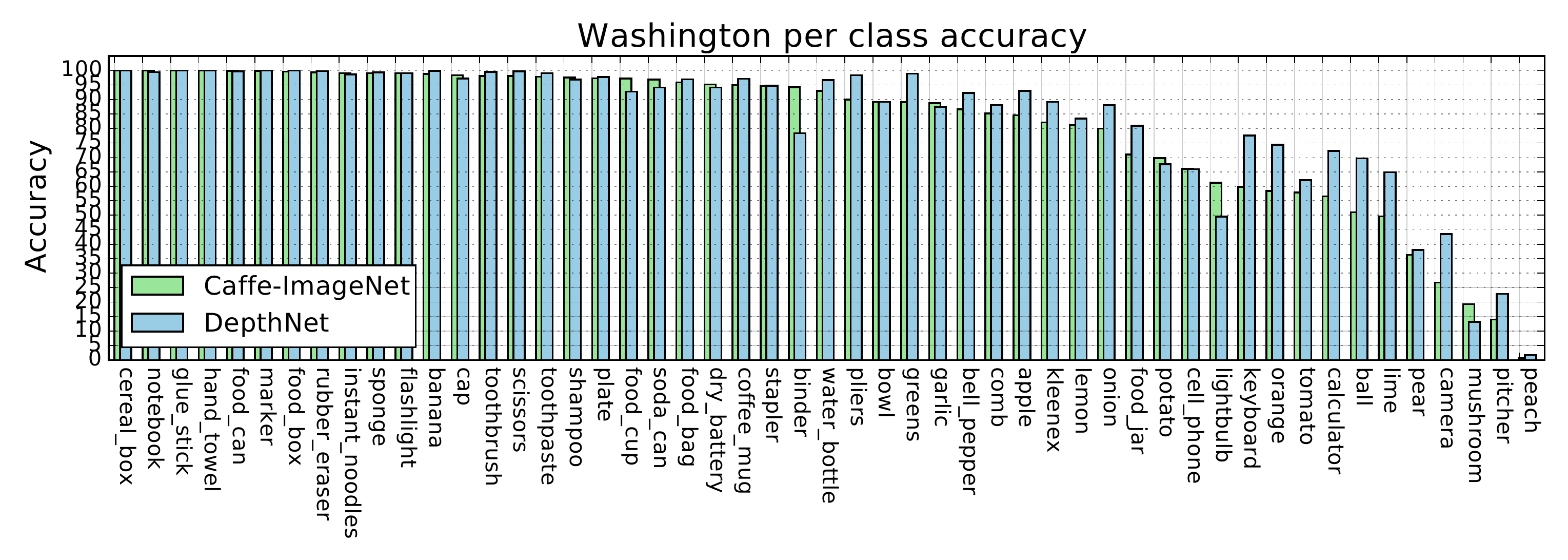}
}
\caption{Accuracy per class on the Washington dataset, depth images. Classes sorted by the  Caffe-ImageNet accuracies.}
\label{fig:washington_acc}
\end{figure*}

Figure \ref{fig:washington_ablation} shows results obtained on the Washington database, with normalized and raw depth data, using as features the activations of the fifth pooling layer (pool5), of the sixth fully connected layer (FC6), and of the seventh fully connected layer (FC7). Note that this last set of activations is the standard choice in the literature. We see that for all settings, pool5 achieves the best performance, followed by FC6 and FC7. This seems to confirm recent findings on RGB data \cite{CNNTransfer}, indicating that pool5 activations offer stronger generalization capabilities when used as features, compared to the more popular FC7. The best performance is obtained by DepthNet, pool5 activations over raw depth data, with a $83.8\%$ accuracy. DepthNet achieves also better results compared to Caffe-ImageNet over normalized data. To get a better feeling of how performance varies when using DepthNet or Caffe-ImageNet, we plotted the per-class accuracies obtained using pool5 and raw depth data. We sorted them in descending order according to the Caffe-ImageNet scores (figure \ref{fig:washington_acc}). 

\begin{figure*}[!htb]
\centering
\subfloat[Device classes as seen by Caffe-ImageNet (left) and DepthNet (right)] {%
\includegraphics[width=0.44\textwidth]{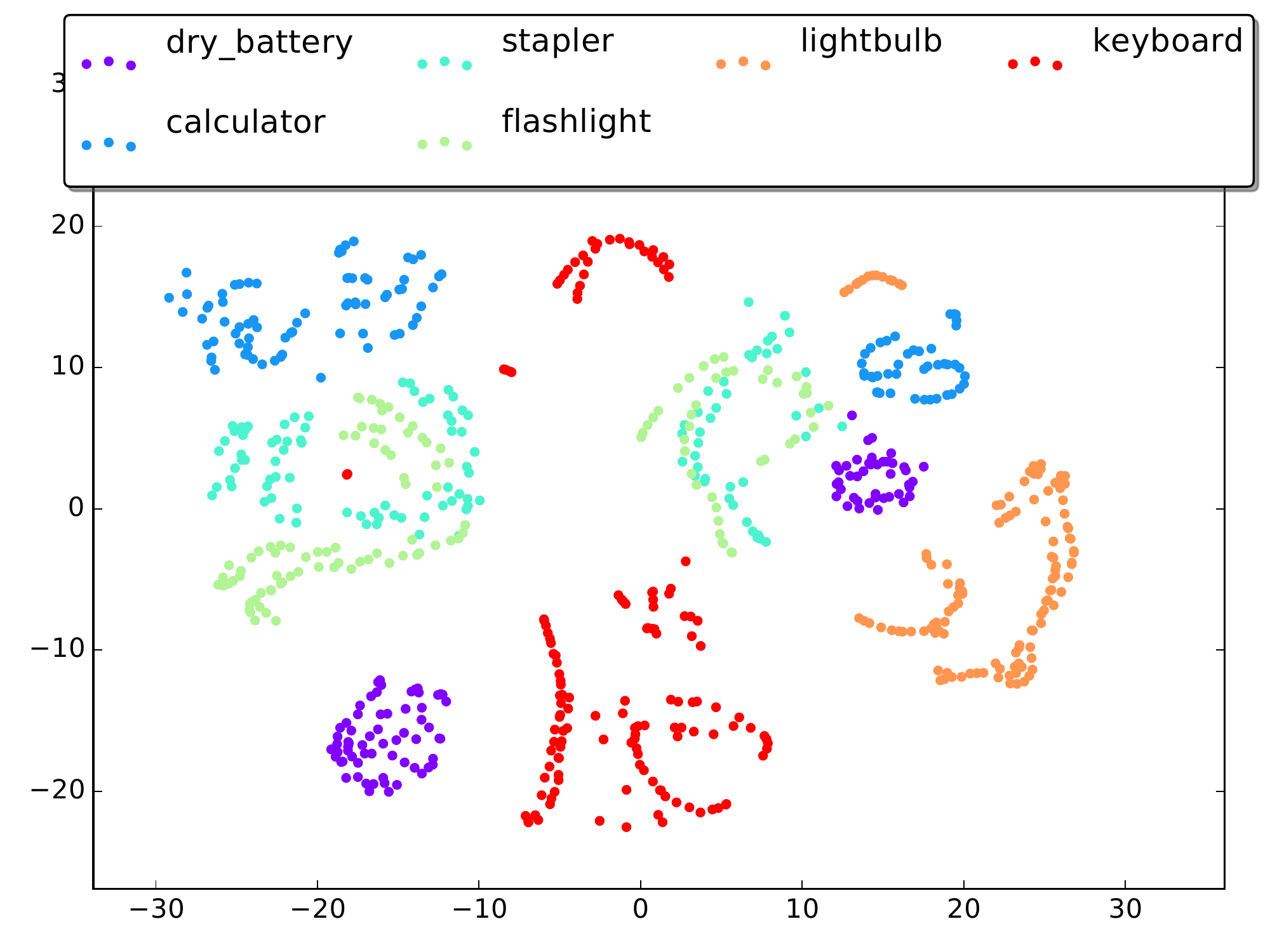}
\includegraphics[width=0.44\textwidth]{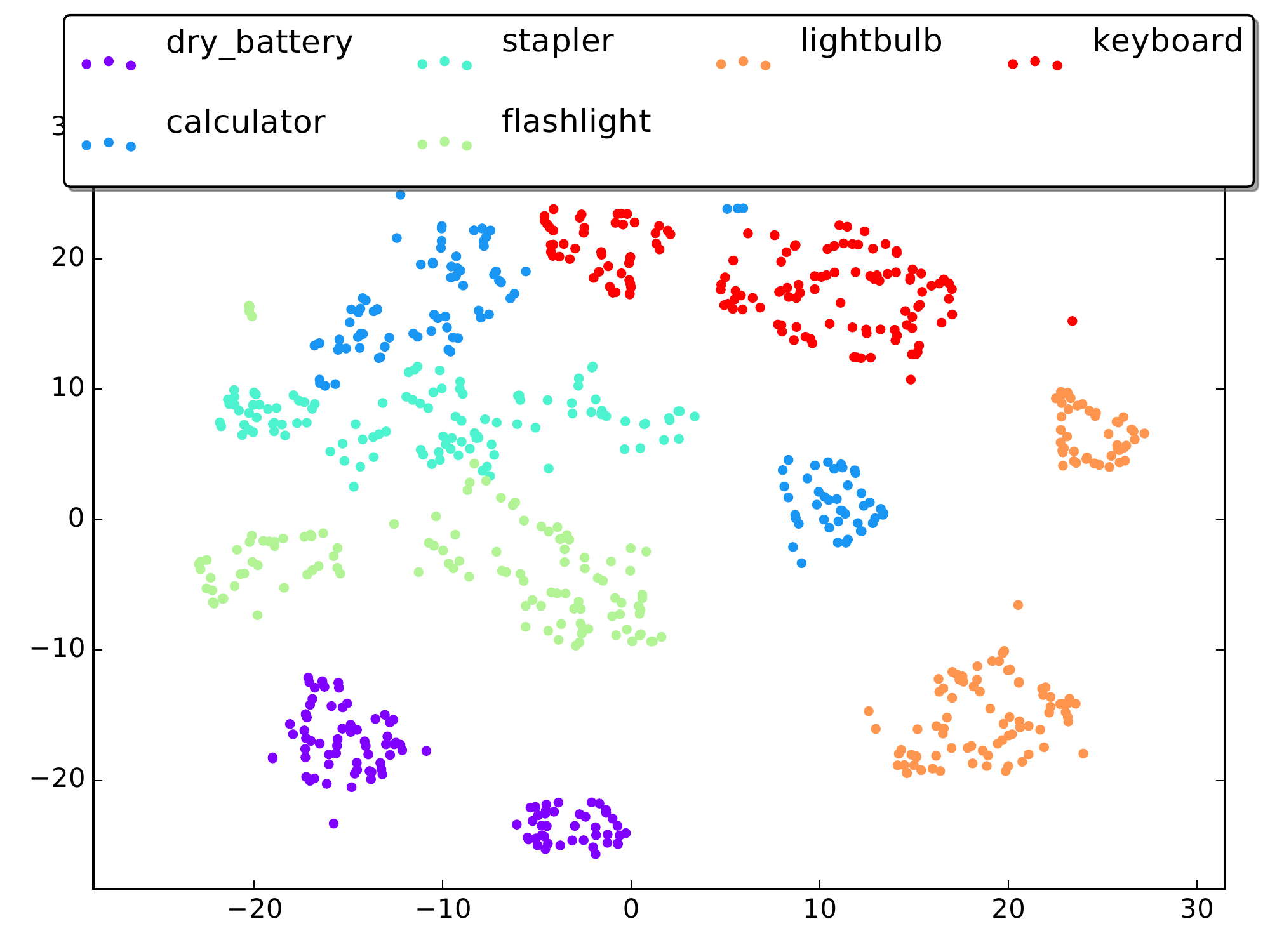}
} \\
\subfloat[Fruit classes as seen by Caffe-ImageNet (left) and DepthNet (right)] {%
\includegraphics[width=0.44\textwidth]{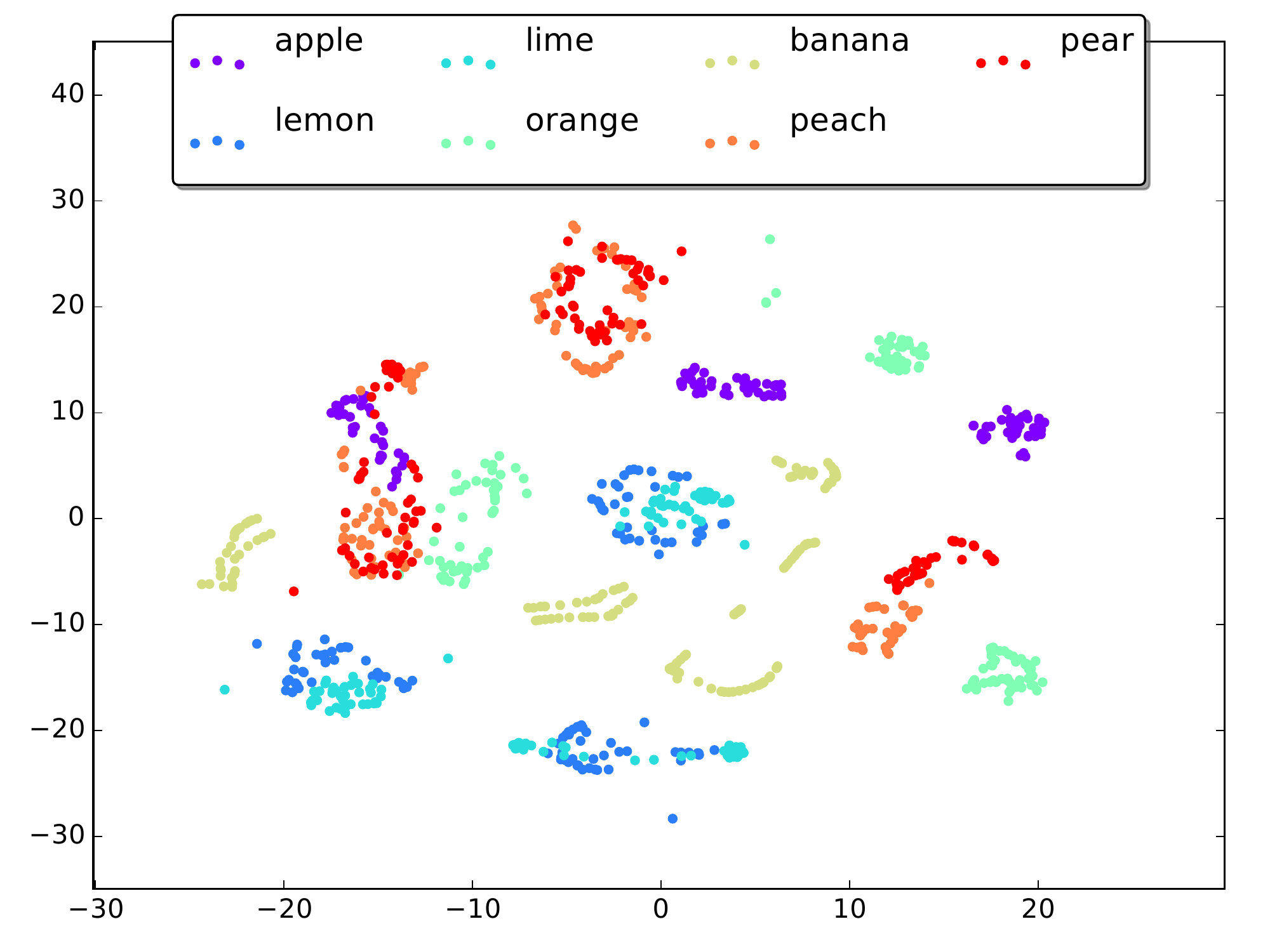}
\includegraphics[width=0.44\textwidth]{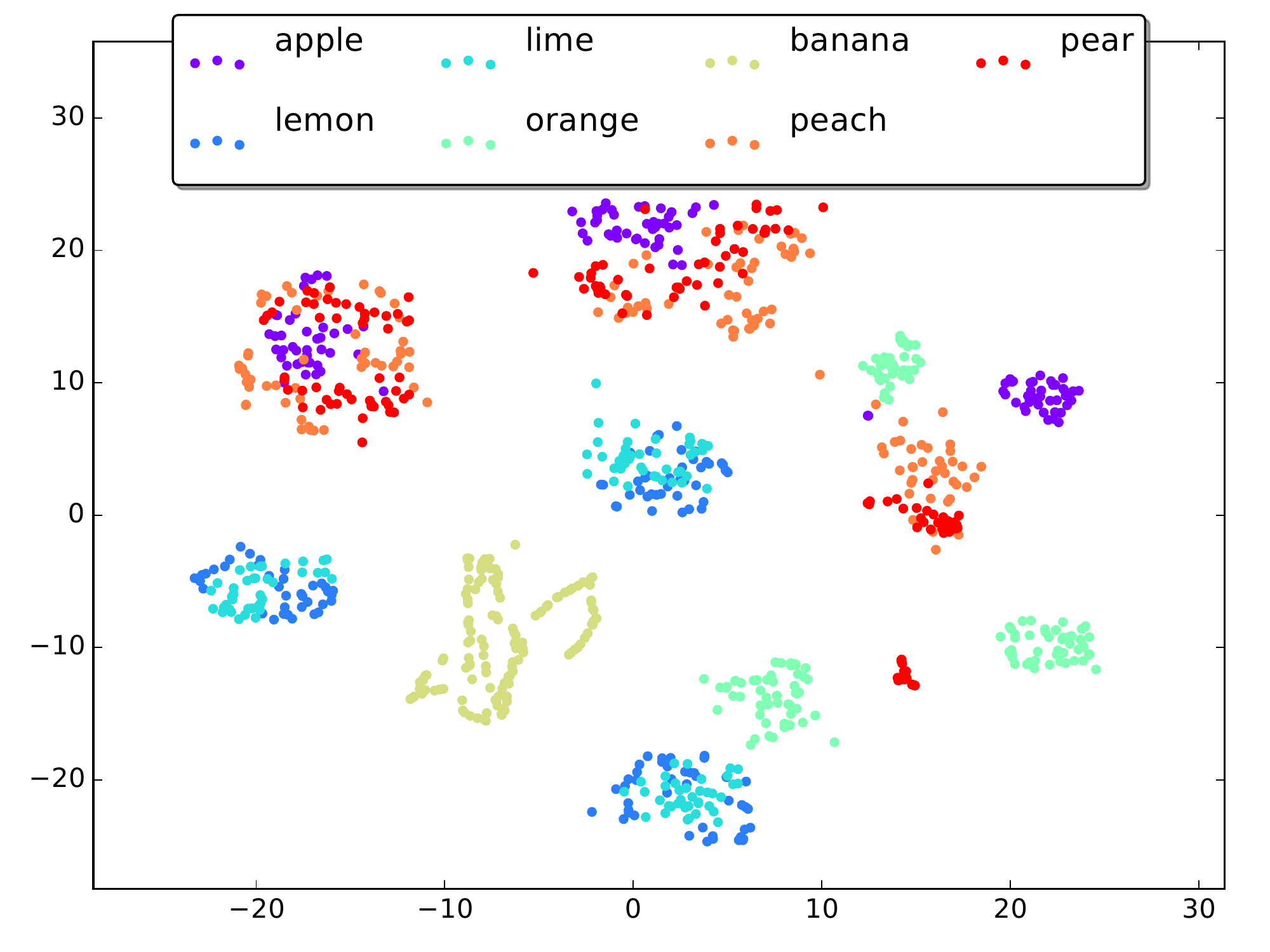}
}
\caption{t-SNE visualizations for the categories device (top) and fruit (bottom).}
\label{fig:features}
\end{figure*}

While there seems to be a bulk of objects where both features perform well (left), DepthNet seems to have an advantage over challenging objects like apple, onion, ball, lime and orange (right), where the round shape tends to be more informative than the specific object texture. This trend is confirmed also when performing a t-SNE visualization \cite{tsne} of all the Washington classes belonging to the high-level categories 'fruit' and 'device' (figure \ref{fig:features}). We see that in general the DepthNet features tend to cluster tightly the single categories while at the same time separating them very well. For some classes like dry battery and banana, the different between the two representations is very marked.
This does not imply that DepthNet features are always better than those computed by Caffe-ImageNet. Figure \ref{fig:washington_acc} shows that CaffeNet features obtain a significantly better performance compared to DepthNet over the classes binder and mushroom, to name just a few. The features learned by the two networks seem to focus on different perceptual aspects of the images. This is most likely due to the different set of samples used during training, and the consequent different filters learned by them (figure \ref{fig:first_layer_weights}).

From these figures we can draw the following conclusions: (a) DepthNet provides the overall stronger descriptor for depth images, regardless of the activation layer chosen and the presence or not of preprocessing on the input depth data; (b) the features derived by the two networks tend to capture different features of the data, and as such are complementary. As we will show in the next section, this last point leads to very strong results when combining the two with a principled cue integration algorithm.

\subsection{Assessing the performance of the RGB-D architecture}
\label{rgb-d}

In this section we present experiments on RGB-D data, from both the Washington and JHUIT databases, assessing the performance of our DepthNet-based framework of figure \ref{fig:architecture} against previous approaches. Table \ref{desp:washington} shows in the top row our results, followed by results obtained by Caffe-ImageNet using the pool5 activations as features, as well as results from the recent literature based on convolutional neural networks. First, we see that the results in the RGB column stresses once more the strength of the pool5 activations as features: they achieve the best performance without any form of fine tuning, spatial pooling or sophisticated non-linear learning, as done instead in other approaches \cite{Eitel,hasan,Cheng}. Second, DepthNet on raw depth data achieves the best performance among CNN-based approaches with or without fine tuning like \cite{Behnke,Eitel}, but it is surpassed by approaches encoding explicitly spatial information through pooling strategies, and/or by using a more advanced classifier than a linear SVM, as we did. We would like to stress that we did not incorporate any of those strategies in our framework on purpose, to better assess the sheer power of training a given convolutional architecture on perceptually different databases. Still, nothing prevents in future work the merging of DepthNet with the best practices in spatial pooling and non-linear classifiers, with a very probable further increase in performance. Lastly, we see that in spite of the lack of such powerful tools, our framework achieves the best performance on RGB-D data. This clearly underlines that the representations learned by DepthNet are both powerful and able to extract different nuances from the data than Caffe-ImageNet.  Rather than the actual overall accuracy reported here in the table, we believe this is the breakthrough result we offer to the community in this paper.

\begin{table*}[!htb]
\centering
$\begin{array}{|l|c|c|c|c|}
\hline
\text{Method:}                                      & \text{RGB}          & \text{Depth Mapping}  & \text{Depth Raw}    & \text{RGB-D}        \\
\hline
\textbf{DepthNet RGB-D Framework}                                           & \mathbf{88.49 \pm 1.8}           & 81.68 \pm 2.2       & \mathbf{83.8 \pm 2.0}  & \mathbf{92.25 \pm 1.3} \\
\hline
\hline
\text{Caffe-ImageNet Pool5}                    & \mathbf{88.49 \pm 1.8} & 81.11 \pm 2  & 78.35 \pm 2.5 & 90.79 \pm 1.2 \\
\text{Caffe-ImageNet FC7 finetuning\cite{Eitel}} & 84.1 \pm 2.7  & 83.8 \pm 2.7 & -            & 91.3 \pm 1.4  \\
\text{Caffe-ImageNet FC7\cite{Behnke}}              & 83.1 \pm 2.0  & -           & -            & 89.4 \pm 1.3 \\
\text{CNN only\cite{Cheng}}                     &     82.7 \pm 1.2 &  78.1 \pm 1.3 & - & 87.5 \pm 1.1 \\
\text{CNN + FisherKernel + SPM\cite{Cheng}}     &     86.8 \pm 2.2 &  \mathbf{85.8 \pm 2.3} & - & 91.2 \pm 1.5 \\
\text{CNN + Hypercube Pyramid + EM\cite{hasan}}  & 87.6  \pm 2.2 &  85.0 \pm 2.1 & - & 91.4 \pm 1.4 \\
\text{CNN-SPM-RNN+CT\cite{ChengZhao2015}}       &     85.2 \pm 1.2 &  83.6 \pm 2.3 & - & 90.7 \pm 1.1 \\
\text{CNN-RNN+CT\cite{ChengZhao2014}}         &	  81.8 \pm 1.9 &  77.7 \pm 1.4 & - & 87.2 \pm 1.1 \\
\text{CNN-RNN\cite{Socher2014}}                                  &     80.8 \pm 4.2 &  78.9 \pm 3.8 & - & 86.8 \pm 3.3 \\
\hline
\end{array}$
\caption{Comparison of our DepthNet framework with previous work on the Washington database. With \textit{depth mapping} we mean all types of depth preprocessing used in the literature.}
\label{desp:washington}
\end{table*}

Experiments over the JHUIT database confirms the findings obtained over the Washington collection (table \ref{table:jhuit-results}). Here our RGB-D framework obtains the second best result, with the state of the art achieved by the proposers of the database with a non CNN-based approach. Note that this database focuses over the fine-grained classification problem, as opposed to object categorization as explored in the experiments above. While the results reported in Table \ref{table:jhuit-results} on Caffe-ImageNet using FC7 seem to indicate that the choice of using pool5 remains valid, the explicit encoding of local information is very important for this kind of tasks \cite{darrell2012,long2014}. We are inclined to attribute to this the superior performance of \cite{jhuit}; future work incorporating spatial pooling in our framework, as well as further experiments on the object identification task in the Washington database and on other RGB-D data collections will explore this issue.




\section{CONCLUSIONS}
\label{conclu}
In this paper we focused on object classification from depth images using convolutional neural networks. We argued that, as effective as the filters learned from ImageNet are, the perceptual features of 2.5D images are different, and that it would be desirable to have deep architectures able to capture them. To this purpose, we created VANDAL, the first depth image database synthetically generated, and we showed experimentally that the features derived from such data, using the very same CaffeNet architecture widely used over ImageNet, are stronger while at the same time complementary to them. This result, together with the public release of the database, the trained architecture and the protocol for generating new depth synthetic images, is the contribution of this paper.
\begin{table}[!tb]
$\begin{array}{|l|c|c|c|c|}
\hline
\text{Method:}                                      & \text{RGB}          & \text{Depth Mapp.}  & \text{Depth Raw}    & \text{RGB-D}        \\
\hline
\textbf{DepthNet Pool5}                                           & -            & \mathbf{54.37}       & \mathbf{55.0}  & \mathbf{90.3} \\
\text{Caffe-ImageNet Pool5}                      & 88.05 & 53.6  & 38.9 & 89.6 \\
\text{Caffe-ImageNet FC7\cite{Behnke}}         &  82.08 & 47.87 & 26.11 & 83.6 \\
\hline
\text{CSHOT + Color pooling} & & & & \\
\text{+ MultiScale Filters\cite{jhuit}} &  - & - & - & \mathbf{91.2} \\
\text{HMP\cite{jhuit}}                                          & 81.4 & 41.1 & - & 74.6 \\
\hline
\end{array}$
\caption{Comparison of our DepthNet framework with previous work on the JHUIT database.
As only one split is defined, we do not report std.}
\label{table:jhuit-results}
\end{table}

We see this work as the very beginning of a long research thread. By its very nature, DepthNet could be plugged into all previous work using CNNs pre-trained over ImageNet for extracting depth features. It might substitute that module, or it might complement it; the open issue is when this will prove beneficial in terms of spatial pooling approaches, learning methods and classification problems. A second issue we plan to investigate is the impact of the deep architecture over the filters learned from VANDAL. While in this work we chose on purpose to not deviate from CaffeNet, it is not clear that this architecture, which was heavily optimized over ImageNet, is able to exploit at best our synthetic depth database. While preliminary investigations with existing architectures have not been satisfactory, we believe that architecture surgery might lead to better results. Finally, we believe that the possibility to use synthetic data as a proxy for real images opens up a wide array of possibilities: for instance, given prior knowledge about the classification task of interest, would it be possible to generate on the fly a task specific synthetic database, containing the object categories of interest under very similar imaging conditions, and train and end-to-end deep network on it? How would performance change compared to the use of network activations as done today? Future work will focus on these issues.


\end{document}